\documentclass{article}
\usepackage[main,final]{neurips_2026}

\usepackage[table]{xcolor}
\usepackage[utf8]{inputenc}
\usepackage[T1]{fontenc}
\usepackage{hyperref}
\usepackage{url}
\usepackage{booktabs}
\usepackage{amsfonts}
\usepackage{nicefrac}
\usepackage{microtype}
\usepackage{xcolor}
\usepackage{graphicx}
\usepackage{subcaption}
\usepackage{caption}
\usepackage{array}
\usepackage{multirow}
\usepackage{makecell}
\usepackage{listings}
\usepackage{threeparttable}
\usepackage{siunitx}
\usepackage{adjustbox}
\usepackage{amsmath}
\usepackage{amssymb}
\usepackage{mathtools}
\usepackage{amsthm}
\usepackage{xspace}
\usepackage[capitalize,noabbrev]{cleveref}

\theoremstyle{plain}

\theoremstyle{definition}

\theoremstyle{remark}

\newcommand{\ours}{\textsc{Mediator}\xspace}

\title{Discrete Neural Interlingua for Ontology-Agnostic EHR Modeling}
\author{%
  Tong Ding$^{1,2\dagger}$ \enspace Caiwei Tian$^{1,2\dagger}$ \enspace Dongmin Bang$^{1,2}$ \enspace Ming Y. Lu$^{1,2}$ \enspace Sophia J. Wagner$^{1,2}$ \\
  \textbf{Andrew Zhang$^{1,2,3}$ \enspace Rowland Pettit$^{1,2}$ \enspace Long Phi Le$^{1,2}$ \enspace Faisal Mahmood$^{1,2}$} \\
  $^1$Mass General Brigham AI Institute \quad $^2$Harvard University \\ $^3$Massachusetts Institute of Technology
}

\begin{document}

\maketitle
{\renewcommand{\thefootnote}{$\dagger$}\footnotetext{Equal contribution.}}

\begin{abstract}
EHR foundation models promise scalable clinical prediction across health systems, yet their deployment is constrained by a tokenization bottleneck. The same clinical concept can appear as different ontology codes, local identifiers, synonyms, or free-text descriptions, which standard EHR tokenizers treat as unrelated symbols. A model trained on MIMIC-III with a closed-vocabulary tokenizer can use only 13\% of eICU events. We introduce \ours, a discrete ontology-agnostic tokenizer that maps any clinical concept expressed as text into a shared vector-quantized codebook, turning EHR harmonization into open-vocabulary tokenization without manual mappings or a fixed vocabulary. Trained with synonym supervision and guided by enriched concept descriptions, code co-occurrence, and relation graphs, \ours collapses equivalent expressions while organizing distinct concepts in a medically meaningful discrete space. Used as the token interface of an EHR Transformer, \ours improves mean AUPRC over the strongest prior tokenizer by 4.0 points (8\% relative) on 20 MIMIC-III tasks and by 8.6 points (29\% relative) in zero-shot transfer to 14 eICU tasks, where it uses all events. The transfer gain combines coverage and representation, as \ours remains the best tokenizer under identical inputs (+1.0 and +1.9 points on operational and subcohort-mortality tasks). Same-encoder controls show that the text encoder drives in-domain accuracy, while the learned objectives of \ours drive transfer robustness and entity-linking quality. \ours also enables transferable attribution across hospitals.
\end{abstract}

\section{Introduction}

\begin{figure}[t!]
  \begin{center}
    \centerline{\includegraphics[width=\columnwidth]{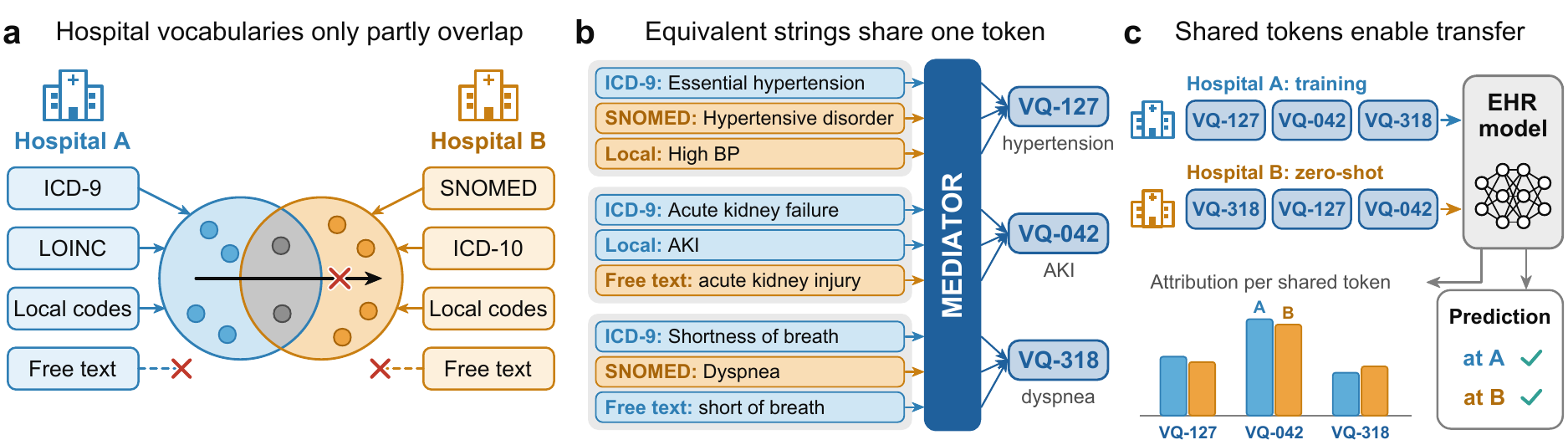}}
    \caption{\textbf{Motivation for \ours.}
\textbf{a}. The vocabularies of Hospital A (light blue) and Hospital B (orange) only partly overlap, so models trained at A fail on B-specific codes and drop free text.
\textbf{b}. \ours maps equivalent surface forms from both hospitals, colored by source hospital, to one shared vector-quantized (VQ) token (dark blue).
\textbf{c}. Shared tokens let one EHR model transfer across hospitals, with predictions and attributions defined over the same tokens.}
    \vspace{-30pt}
    \label{fig:hook}
  \end{center}
\end{figure}

Electronic health records (EHRs) are central to clinical prediction, real-world evidence generation, cohort construction, and decision support. Transformer-based EHR models have achieved strong results by learning from longitudinal sequences of diagnoses, procedures, medications, laboratory events, and other clinical observations \citep{delphi, comet, li2020behrt, ehrshot, clmbr}. These models promise to support multi-institutional learning and deployment across health systems. Yet this promise is constrained by a representation problem. Clinical concepts are recorded differently across coding systems, ontology versions, local institutional practice, and free-text documentation. A model trained on MIMIC-III \citep{PhysioNet-mimiciii-1.4} with a closed-vocabulary tokenizer can use only 13\% of eICU events \citep{eicu}, since the rest use other codes or free text. As a result, many EHR models learn from an input space that is fragmented, redundant, and difficult to transfer across settings (\cref{fig:hook}).

Most existing EHR foundation models tokenize structured data by treating each medical code as an independent symbol with a learned embedding \citep{shaky}. This design scales poorly to large, heavy-tailed clinical vocabularies and fails to share statistical strength across related concepts \citep{medtok}. Code-level tokens also discard informative structure. Each code is defined by a textual description, belongs to an ontology hierarchy, and is linked to many synonyms and near-synonyms across other coding systems \citep{clinvec}. Ignoring these semantics leads to inefficient vocabularies and fragmented representations, especially for rare or institution-specific codes. This brittleness is most visible across institutions. Hospitals differ in the ontologies they use, which versions they deploy, and how they encode local variants of common concepts. When a model is trained with one hospital's code vocabulary, deploying it elsewhere typically requires manual code mapping, vocabulary alignment, and coverage audits. Such mapping is expensive, error-prone, and often incomplete in the long tail. As a result, most models avoid clinical text entirely \citep{shaky} and restrict modeling to a narrow subset of structured data such as diagnostic codes, labs, and medications, sacrificing clinical signal for modeling convenience.

Previous studies have attempted to address this heterogeneity from several complementary angles. Ontology-normalization approaches map events to a common vocabulary such as the Unified Medical Language System (UMLS) \citep{umls,takecare}; representation-learning approaches enrich fixed code vocabularies with text descriptions and medical knowledge graphs, and LLM-generated definitions \citep{medtok,medrep}; text-based embedding methods represent codes directly in continuous semantic space \citep{descemb}; and multi-institutional harmonization frameworks align site-specific vocabularies using EHR summary statistics, biomedical graphs, and language-model semantics \citep{zhou2026representation}. These efforts improve semantic coverage and reduce reliance on manual pairwise code mapping, but they do not fully address the modeling interface required for scalable EHR sequence modeling. First, UMLS defines a massive and evolving vocabulary. It aggregates multiple source vocabularies, introducing redundancy and fine-grained distinctions that are often unnecessary for downstream prediction. Consequently, using UMLS identifiers directly as tokens reintroduces the problems of scale, sparsity, and redundancy inherent in raw code vocabularies \citep{medtok}. Fixed-vocabulary code embeddings remain closed over a prespecified set of structured codes, and site-alignment methods generally must be rerun when a new hospital, ontology, or local code system is introduced. Continuous embeddings are more flexible, but they do not provide the stable discrete interface used by standard masked-token and autoregressive EHR Transformers, which also complicate event-level inspection, attribution, caching, and generative trajectory modeling. A useful EHR tokenizer should therefore be both discrete and open-vocabulary. It should accept any medical concept expressed as text and map it into a shared, reusable code space.

To this end, we introduce MEDical InterlinguA TokenizeR (\ours), a discrete and ontology-agnostic concept space that serves as a neural interlingua. \ours takes as input any clinical concept expressible as text (including ontology code descriptions, local code descriptions, and uncoded text entries) and maps it to a discrete index from a fixed-size vector-quantized (VQ) codebook, in which each code is a semantic prototype with an embedding.
Leveraging strong pretrained text encoders, we learn this mapping from large-scale cross-ontology synonym supervision, treating each underlying concept as an equivalence class of surface forms. 
To better shape the representation geometry, we further guide codebook learning with three complementary sources of medical knowledge including enriched concept descriptions, empirical code co-occurrence, and curated relation-graph structure. The resulting codebook defines a compact, reusable clinical vocabulary that can be used as a drop-in token interface for standard EHR Transformers, improving interoperability and interpretability while avoiding fragile ontology-specific tokenization and manual code mapping.

We show that \ours is effective in settings that stress current EHR modeling. As the token interface of an EHR Transformer, \ours improves mean AUPRC over the strongest prior tokenizer by 4.0 points (8\% relative) on 20 MIMIC-III tasks and by 8.6 points (29\% relative) in zero-shot transfer to 14 eICU tasks without manual vocabulary alignment, and remains the best tokenizer when input coverage is matched to closed-vocabulary baselines. Controlled analyses decompose these gains. The transfer gain combines broader input coverage with better representations, the pretrained text encoder drives in-domain accuracy, and the learned objectives of \ours drive transfer robustness and entity-linking quality. Finally, the shared discrete space supports transferable interpretability across hospitals through common concept tokens.

\section{Related work}

\textbf{EHR models.} Transformer-based EHR models adapt the language-modeling paradigm to longitudinal sequences of structured clinical events. Bidirectional embedding models learn patient representations with masked-event or contextual prediction objectives, starting from diagnosis-focused token sequences and later expanding to medications, procedures, temporal features, and discretized numeric observations \citep{li2020behrt,medbert,cehrbert,ehrshot}, and graph-based variants model intra-visit structure \citep{poulain2024gtbehrt}. Autoregressive and encoder--decoder models use generative objectives to model patient trajectories and support forecasting, time-to-event prediction, and generative medical-event modeling \citep{clmbr,motor,foresight,delphi,comet,yang2023transformehr}, and toolkits such as PyHealth 2.0 improve their reproducibility \citep{pyhealth}. These studies underline the importance of tokenized EHR data, but concept embeddings learned from scratch or initialized from external resources remain tied to a closed vocabulary and transfer poorly across institutions.

\textbf{Cross-hospital transfer.} Improving transfer across hospitals requires resolving vocabulary mismatch from different ontologies, local code systems, and documentation practices. One strategy is to normalize structured events to a shared vocabulary before model training, such as UMLS \citep{umls,takecare} or the standard concepts of the OMOP common data model maintained by OHDSI \citep{overhage2012omop,hripcsak2015ohdsi}, which are largely mapped by hand. This inherits the scale, redundancy, and incomplete coverage of the target vocabulary and requires ongoing curation. A second strategy learns continuous code representations from textual descriptions, or enriches a fixed structured vocabulary with biomedical knowledge graphs and LLM-generated concept descriptions \citep{descemb,medtok,medrep}. These representations improve semantic sharing among known codes, but they remain tied to a prespecified vocabulary or yield continuous event vectors that lack a discrete token interface. A third strategy aligns vocabularies across institutions using site-level summary statistics, biomedical graphs, and language-model semantics \citep{zhou2026representation}. However, the alignment is specific to the participating institutions and must be recomputed for new hospitals or unseen code systems. \ours learns an open-vocabulary text-to-token mapping that complements manual OMOP mapping. Any clinical concept expressible as text can be mapped into a shared discrete codebook and then reused by standard EHR sequence models.

\textbf{Vector-quantized tokenization.} Vector-quantized autoencoders such as VQ-VAE \citep{van2017neural} learn a finite codebook of latent prototypes and discretize inputs by nearest-neighbor assignment. Such tokenizers now link perceptual encoders and sequence models in vision, audio, and multimodal learning \citep{esser2021taming, zeghidour2021soundstream, du2024on}. In healthcare, VQ-based discretization has been used for time-series tokenization and ontology-aware token learning over structured medical codes \citep{lee2024leveraging, medtok}, but these tokenizers are trained on a closed set of observed codes. \ours provides an open-vocabulary text-to-token interface that maps unseen code descriptions or free-text concept expressions into the same discrete interlingua.
\section{Methods}

\subsection{Problem definition}

\begin{figure*}[t!]
  \centering
  \includegraphics[width=\textwidth]{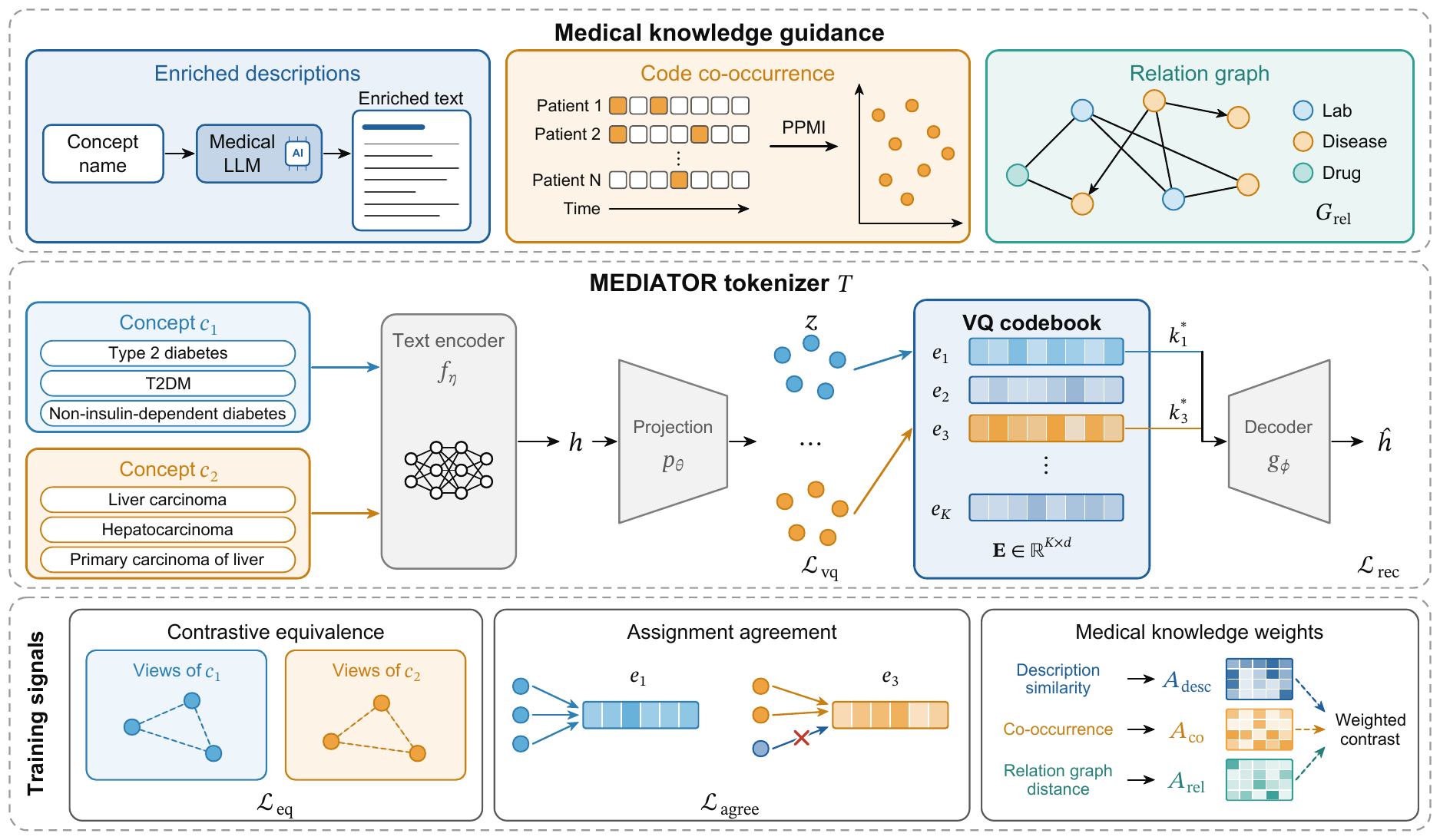}
  \caption{\textbf{\ours architecture.} A text encoder maps any clinical concept description to a dense embedding, a trainable adaptor projects it into the quantization space, and the result is assigned to a learned VQ codebook. Training combines same-concept contrastive objectives before and after quantization, multi-view assignment agreement, semantic teachers from enriched descriptions, code co-occurrence, and relation graphs, and reconstruction to the encoder embedding space.}
  \label{fig:fig2}
  \vspace{-12pt}
\end{figure*}

Let $\mathcal{C}$ be a universe of underlying clinical concepts. Each concept
$c\in\mathcal{C}$ can be expressed by multiple textual descriptions, including
ontology descriptions, local code descriptions, synonyms, and uncoded text
entries. We model these views as $\mathcal{X}_c=\{x_{c,1},\ldots,x_{c,|\mathcal{X}_c|}\}\subset\mathcal{X}$, where $\mathcal{X}$ is the space of text strings. Our goal is to learn a
discrete tokenizer $T:\mathcal{X}\to[K]=\{1,\ldots,K\}$ such that descriptions of the same concept map to the same discrete code, while
different concepts remain separated as much as possible. Figure~\ref{fig:fig2} gives an overview of the tokenizer (middle panel), its training signals (bottom panel), and the medical knowledge that guides training (top panel).

\subsection{Vector-quantized tokenization}

Given $x\in\mathcal{X}$, a text encoder produces a semantic embedding
$h=f_{\eta}(x)\in\mathbb{R}^{d_e}$, and a projection head maps it to the
quantization space $z=p_{\theta}(h)\in\mathbb{R}^{d}$. We maintain a codebook
$\mathbf{E}=\{e_k\}_{k=1}^{K}$ with $e_k\in\mathbb{R}^{d}$. Using cosine
similarity, the assigned code is
\begin{equation}
k^\star(z)=\arg\max_{k\in[K]}\cos(z,e_k),
\qquad
q(z)=e_{k^\star(z)}.
\end{equation}
The tokenizer output is the discrete index
$T(x)=k^\star(p_{\theta}(f_{\eta}(x)))$.

We update codebook statistics using exponential moving averages (EMA), following
standard VQ practice~\citep{van2017neural,razavi2019generating}. Let $n_k$ be
the number of assignments to code $k$ in a minibatch and
$m_k=\sum_{i:k^\star(z_i)=k}z_i$ be the sum of assigned vectors. We maintain EMA statistics $(N_k,M_k)$ and update
\begin{equation}
N_k \leftarrow \gamma N_k+(1-\gamma)n_k,\qquad
M_k \leftarrow \gamma M_k+(1-\gamma)m_k,\qquad
e_k \leftarrow M_k/(N_k+\varepsilon),
\end{equation}
with decay $\gamma\in(0,1)$ and small $\varepsilon>0$. We use the commitment
objective $\mathcal{L}_{\mathrm{vq}}=\frac{1}{N}\sum_{i=1}^{N}\|z_i-\mathrm{sg}(q(z_i))\|_2^2$, where $\mathrm{sg}(\cdot)$ stops gradients.

To avoid permanent
codebook death, we periodically reinitialize under-used codes from the empirical
distribution of current encoder outputs, following dead-code revival in neural
audio tokenization~\citep{zeghidour2021soundstream}. For losses applied after
quantization, we use the rotation trick~\citep{fifty2025restructuring}. Let $\hat z$ denote the surrogate
quantized representation used for backpropagation. Its forward value is
$q(z)$, while its approximate Jacobian rotates gradients as
\begin{equation}
\frac{\partial \hat z}{\partial z}\approx \alpha R(z,q(z)),
\qquad
\nabla_z\mathcal{L}\approx
\alpha R(z,q(z))^{\top}\nabla_{\hat z}\mathcal{L},
\end{equation}
where $R(\cdot,\cdot)$ is an orthonormal map from the direction of $z$ to
$q(z)$ and $\alpha>0$ rescales magnitudes.

Finally, a decoder $g_{\phi}$ reconstructs the encoder embedding with $\mathcal{L}_{\mathrm{rec}}=1-\frac{1}{N}\sum_{i=1}^{N}\cos(g_{\phi}(\hat z_i),h_i)$ to preserve clinically relevant semantics while
discouraging code collapse~\citep{qu2025tokenflow}.

\subsection{Equivalence supervision from multiple views}

The primary supervision signal is concept equivalence. All views of the same
underlying concept should be interchangeable at the tokenizer interface. Given
a minibatch of $N$ views with concept labels $c_i$, let
$\mathcal{P}(i)=\{j\neq i:c_j=c_i\}$
be the set of other views of the same concept. For a representation $r_i$,
define
\begin{equation}
\pi_{ij}(r)
=\frac{\exp(\cos(r_i,r_j)/\tau)}
{\sum_{\ell\neq i}\exp(\cos(r_i,r_{\ell})/\tau)}.
\end{equation}
We use a multi-positive contrastive loss
\begin{equation}
\mathcal{L}_{\mathrm{eq}}(r)
=-\frac{1}{|\mathcal{I}|}\sum_{i\in\mathcal{I}}
\log \sum_{j\in\mathcal{P}(i)}\pi_{ij}(r),
\end{equation}
where $\mathcal{I}=\{i:|\mathcal{P}(i)|>0\}$. This loss is applied before
VQ to organize the continuous space, $\mathcal{L}_{\mathrm{eq}}^{\mathrm{pre}}=\mathcal{L}_{\mathrm{eq}}(\psi_{\mathrm{pre}}(z))$, and after quantization to act directly on the discrete interface, $\mathcal{L}_{\mathrm{eq}}^{\mathrm{post}}=\mathcal{L}_{\mathrm{eq}}(\psi_{\mathrm{post}}(\hat z))$.

We further align the assignment distributions of views from the same concept.
Let $p_i(k)=\exp(\cos(z_i,e_k)/\tau_a)/\sum_{\ell=1}^{K}\exp(\cos(z_i,e_{\ell})/\tau_a)$
be the soft code-assignment distribution. For concept $c$ in a minibatch, let
$\mathcal{V}_c=\{i:c_i=c\}$ and
$\bar p_c=\frac{1}{|\mathcal{V}_c|}
\sum_{i\in\mathcal{V}_c}p_i.$
The assignment-agreement loss, a generalized Jensen--Shannon objective over code assignments, is
\begin{equation}
\mathcal{L}_{\mathrm{agree}}
=\frac{1}{|\mathcal{C}_B|}
\sum_{c\in\mathcal{C}_B}
\frac{1}{|\mathcal{V}_c|}
\sum_{i\in\mathcal{V}_c}
\mathrm{KL}(p_i\,\|\,\bar p_c),
\end{equation}
where $\mathcal{C}_B$ is the set of concepts represented in the minibatch. 

\subsection{Medical knowledge as semantic teachers}

The discrete tokenizer is the desired interface, but learning a high-quality
codebook benefits from a well-structured representation space before the VQ
bottleneck. We therefore use medical knowledge to define soft semantic teachers
over non-equivalent concepts. 

We use three categories of medical knowledge. First, \emph{LLM-generated enriched
descriptions} provide definitional
knowledge about what a concept means. Let $y_c$ denote an enriched description
for concept $c$ and let
$u_c=\rho(y_c)/\|\rho(y_c)\|_2$ be its normalized embedding. We define
\begin{equation}
A_{\mathrm{desc}}(c_i,c_j)
=\exp\!\left(\frac{\max(0,u_{c_i}^{\top}u_{c_j})}{\tau_{\mathrm{desc}}}\right)-1.
\end{equation}
Second, \emph{code co-occurrence} provides real-world evidence about which
concepts appear in similar clinical contexts. We define co-occurrence over a
set of contexts, where each context corresponds to a patient and a fixed-width
time window. Let $B\in\{0,1\}^{D\times |\mathcal{C}|}$ be the binary
context--concept matrix, where $B_{tc}=1$ if concept $c$ appears in context
$t$. The sparse co-occurrence matrix is $C=B^{\top}B$, and the marginal context count for concept $i$ is $n_i=C_{ii}$. We form a raw
positive pointwise mutual information (PPMI) matrix
\begin{equation}
M_{ij}=
\begin{cases}
\max\!\left(\log_2\frac{D\,C_{ij}}{n_i n_j},\,0\right),
& C_{ij}>0,\\
0, & C_{ij}=0.
\end{cases}
\end{equation}
We define $A_{\mathrm{co}}(c_i,c_j)=\log(1+M_{c_i c_j})/\tau_{\mathrm{co}}$, which preserves empirical usage evidence while damping very large PPMI values. Third, the \emph{relation graph} captures curated medical relations. Given a graph $G_{\mathrm{rel}}$ over
concepts and shortest-path distance $d_{\mathrm{rel}}$, we define
\begin{equation}
A_{\mathrm{rel}}(c_i,c_j)
=
\begin{cases}
\exp(-d_{\mathrm{rel}}(c_i,c_j)/\tau_{\mathrm{rel}}), &
d_{\mathrm{rel}}(c_i,c_j)\le R,\\
0, & \text{otherwise}.
\end{cases}
\end{equation}

Let $\mathcal{M}=\{\mathrm{desc},\mathrm{co},\mathrm{rel}\}$. For each teacher
$m\in\mathcal{M}$, we convert affinities in a minibatch into a row-normalized
target distribution over non-equivalent concepts
\begin{equation}
\tilde w_{ij}^{(m)}
=\frac{\mathbf{1}[c_i\neq c_j]A_m(c_i,c_j)}
{\sum_{\ell\neq i}\mathbf{1}[c_i\neq c_{\ell}]A_m(c_i,c_{\ell})}.
\end{equation}
Rows with zero total affinity are omitted, and sparse teachers keep only the
largest affinities per row. With
$\pi_{ij}^{\mathrm{pre}}=\pi_{ij}(\psi_{\mathrm{pre}}(z))$, the teacher loss is
\begin{equation}
\mathcal{L}_{m}
=-\frac{1}{|\mathcal{I}_m|}
\sum_{i\in\mathcal{I}_m}
\sum_{j\neq i}\tilde w_{ij}^{(m)}\log \pi_{ij}^{\mathrm{pre}}.
\end{equation}
Same-concept objectives determine what must collapse, and the semantic teachers determine how distinct concepts should be arranged before quantization.

\subsection{Training objective and inference}

The final objective, combining vector quantization, semantic reconstruction,
same-concept equivalence, assignment agreement, and the semantic teachers, is
\begin{equation}
\mathcal{L}
=\lambda_{\mathrm{pre}}\mathcal{L}_{\mathrm{eq}}^{\mathrm{pre}}
+\lambda_{\mathrm{post}}\mathcal{L}_{\mathrm{eq}}^{\mathrm{post}}+\lambda_{\mathrm{vq}}\mathcal{L}_{\mathrm{vq}}
+\lambda_{\mathrm{rec}}\mathcal{L}_{\mathrm{rec}}+\lambda_{\mathrm{agree}}\mathcal{L}_{\mathrm{agree}}
+\sum_{m\in\mathcal{M}}\lambda_{m}\mathcal{L}_m .
\end{equation}

At inference time, \ours is applied as an offline preprocessing step to unique clinical entity strings, and the resulting
discrete token IDs $T(x)$ are cached and reused for all patient sequences. After this one-time tokenization step, downstream EHR models operate on discrete token IDs in the same way as standard code-embedding models. Because the codebook is fixed, new code systems such as ICD-11 can be tokenized without retraining, as in our eICU and entity-linking experiments (Appendix~\ref{app:worked_example}); novel clinical distinctions remain untested.
\section{Experiments}

\begin{figure*}[t!]
  \begin{center}
    \centerline{\includegraphics[width=\textwidth]{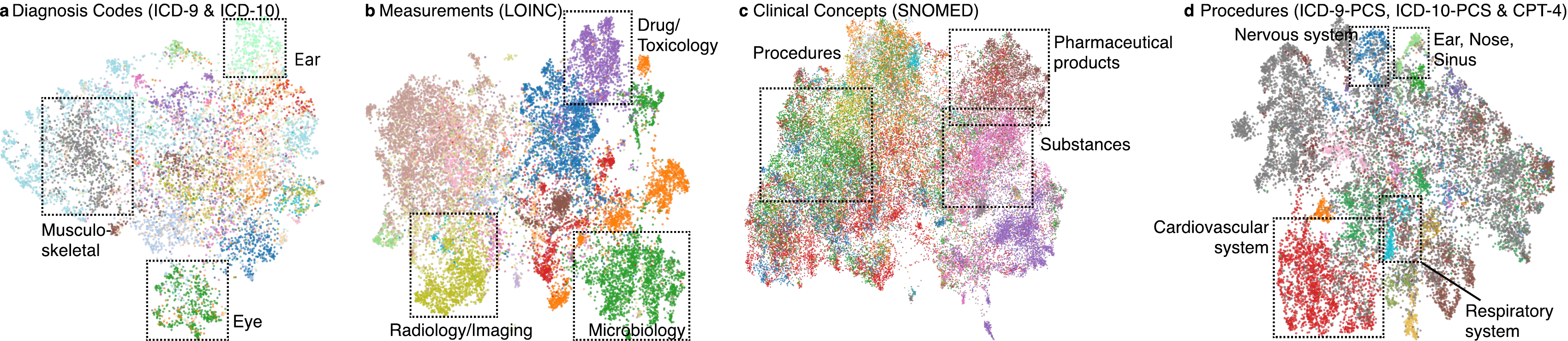}}
    \caption{\textbf{UMAP visualization of the \ours codebook.}
    Each point is a source concept, shown by its assigned VQ code embedding projected with UMAP. Colors indicate independent ontology-derived categories for (a) diagnosis codes, (b) measurements, (c) SNOMED clinical concepts, and (d) procedures. Dashed boxes mark example categories. Full category legends are in Appendix~\ref{app:umap}.}
    \vspace{-20pt}
    \label{fig:umap}
  \end{center}
\end{figure*}

\definecolor{lightgray}{gray}{0.9}
\begin{table*}[t]
\centering
\caption{EHR prediction performance on MIMIC-III, reported as mean (std) AUPRC across folds. Across the 20 tasks, \ours improves over MedTok (one-sided Wilcoxon signed-rank test, $p=9.5\times10^{-7}$). Patient-level bootstrap tests (5{,}000 resamples) show significant gains ($p<0.05$) on 7/20 tasks over MedTok and 17/20 tasks over Code Emb.}
\label{tab:mimic_results}
\setlength{\tabcolsep}{2.2pt}
\footnotesize
\begin{tabular}{lcccccc}
\toprule
Task
  & Code Emb & ClinVec & MedTok
  & $\text{MedTok}_{\text{text}}$-top5
  & $\text{MedTok}_{\text{text}}$-top1
  & \textbf{\ours} \\
\midrule
\multicolumn{7}{l}{\textit{Operational tasks}} \\
LOS $<$7
  & 0.819(0.012) & 0.807(0.004) & 0.827(0.007)
  & \textbf{0.830(0.005)} & 0.762(0.011) & 0.829(0.007)\\
ICU Mortality
  & 0.571(0.015) & 0.484(0.006) & 0.556(0.008)
  & 0.563(0.007) & 0.418(0.014) & \textbf{0.573(0.010)} \\
Readmission
  & 0.336(0.026) & 0.342(0.018) & 0.366(0.021)
  & 0.347(0.021) & 0.336(0.022) & \textbf{0.374(0.023)} \\
\rowcolor{lightgray}
Mean
  & 0.576 & 0.544 & 0.583
  & 0.580 & 0.505 & \textbf{0.592} \\
\midrule
\multicolumn{7}{l}{\textit{Subcohort mortality tasks}} \\
Acute Kidney Injury
  & 0.563(0.032) & 0.513(0.025) & 0.573(0.014)
  & 0.572(0.024) & 0.436(0.013) & \textbf{0.587(0.013)} \\
Acute MI
  & 0.376(0.071) & 0.397(0.037) & 0.412(0.021)
  & 0.404(0.045) & 0.337(0.035) & \textbf{0.467(0.056)} \\
Acute Pancreatitis
  & 0.263(0.053) & 0.307(0.084) & 0.359(0.074)
  & 0.311(0.054) & 0.234(0.061) & \textbf{0.396(0.082)} \\
Atrial Fibrillation
  & 0.489(0.025) & 0.416(0.027) & 0.473(0.041)
  & 0.482(0.023) & 0.383(0.051) & \textbf{0.497(0.012)} \\
ARF/ARDS
  & 0.542(0.028) & 0.520(0.025) & 0.585(0.005)
  & 0.586(0.018) & 0.497(0.027) & \textbf{0.603(0.030)} \\
Cirrhosis
  & 0.291(0.072) & 0.355(0.135) & 0.358(0.150)
  & 0.224(0.098) & 0.195(0.044) & \textbf{0.444(0.120)} \\
Heart Failure
  & 0.426(0.026) & 0.362(0.031) & 0.403(0.030)
  & 0.430(0.026) & 0.305(0.023) & \textbf{0.447(0.050)} \\
Liver Failure
  & 0.564(0.055) & 0.577(0.061) & 0.599(0.029)
  & 0.566(0.083) & 0.470(0.073) & \textbf{0.615(0.024)} \\
GI Bleeding
  & 0.350(0.052) & 0.398(0.049) & 0.470(0.085)
  & 0.430(0.049) & 0.298(0.035) & \textbf{0.479(0.032)} \\
Severe Sepsis/Septic Shock
  & 0.581(0.035) & 0.593(0.018) & 0.640(0.012)
  & 0.634(0.019) & 0.514(0.029) & \textbf{0.676(0.034)} \\
Cerebrovascular Disease
  & 0.514(0.049) & 0.539(0.057) & 0.577(0.014)
  & 0.561(0.031) & 0.525(0.017) & \textbf{0.631(0.038)} \\
Subarachnoid Hemorrhage
  & 0.296(0.064) & 0.317(0.135) & 0.394(0.161)
  & 0.360(0.070) & 0.362(0.119) & \textbf{0.556(0.155)} \\
\rowcolor{lightgray}
Mean
  & 0.438 & 0.441 & 0.487
  & 0.463 & 0.380 & \textbf{0.533} \\
\midrule
\multicolumn{7}{l}{\textit{Survival tasks}} \\
Liver Cancer
  & 0.609(0.058) & 0.611(0.077) & 0.681(0.133)
  & 0.625(0.052) & 0.642(0.062) & \textbf{0.719(0.068)} \\
Prostate Cancer
  & 0.460(0.179) & 0.501(0.131) & 0.557(0.185)
  & 0.371(0.121) & 0.472(0.181) & \textbf{0.584(0.149)} \\
Brain Cancer
  & 0.607(0.055) & 0.558(0.048) & 0.550(0.081)
  & 0.586(0.087) & 0.533(0.101) & \textbf{0.650(0.109)} \\
Myeloid Leukemia
  & 0.570(0.119) & 0.586(0.095) & 0.595(0.109)
  & 0.589(0.085) & 0.531(0.058) & \textbf{0.595(0.056)} \\
Lymphoid Leukemia
  & 0.542(0.144) & 0.547(0.089) & 0.580(0.141)
  & 0.503(0.114) & 0.614(0.056) & \textbf{0.646(0.039)} \\
\rowcolor{lightgray}
Mean
  & 0.557 & 0.561 & 0.593
  & 0.535 & 0.558 & \textbf{0.639} \\
\bottomrule
\end{tabular}
\vspace{-18pt}
\end{table*}

\textbf{\ours pretraining dataset.}
We pretrain \ours on a concept-view corpus derived from UMLS 2025AB \citep{umls}. We first merge CUIs linked by synonym relations into concept groups. All English, unsuppressed names and source descriptions within a group are positive views of the same concept. The corpus spans the International Classification of Diseases (ICD-9 \citep{WHO1988ICD9} and ICD-10 \citep{icd10}), SNOMED CT US \citep{snomed}, LOINC \citep{loinc}, RxNorm \citep{RXNORM}, CPT \citep{cpt}, HCPCS \citep{HCPCS}, CVX \citep{cdc_cvx}, ATC \citep{who_atc_index_2026}, and MeSH \citep{nlm_mesh}. We also align LLM-enriched descriptions, a PPMI matrix from claims-derived code co-occurrence, and a medical relation KG to the same concept groups. 
Details are in Appendix~\ref{vq_training_data}.

\textbf{Biomedical entity linking (BEL) benchmarks.}
We use three curated BEL datasets, NCBI Disease \citep{DOGAN20141}, MedMentions \citep{Mohan2019MedMentionsAL}, and COMETA \citep{basaldella-etal-2020-cometa}, which map medical mentions from scientific text and social media to ontology concepts and are standard benchmarks for synonym alignment \citep{sapbert}. We tokenize each mention and report pairwise consistency, which measures whether synonymous strings collapse to the same code, code purity, which measures semantic coherence within each code, and their harmonic mean $\mathrm{F1} = 2\,\mathrm{Consistency}\cdot\mathrm{Purity}/(\mathrm{Consistency}+\mathrm{Purity})$.

\textbf{EHR benchmark (in-domain).}
We evaluate whether \ours serves as a better tokenizer for downstream EHR sequence modeling on MIMIC-III \citep{PhysioNet-mimiciii-1.4}. We build patient timelines from diagnoses, procedures, medications, and other coded events, restricting inputs to events timestamped at or before the prediction time (diagnoses and procedures, which MIMIC-III records only at discharge, enter only discharge-time tasks; Appendix~\ref{mimic_task}), train the same BEHRT-style Transformer from scratch for each task and tokenizer, and report five-fold mean AUPRC on operational, disease-specific mortality, and survival tasks.

\textbf{Zero-shot cross-dataset generalization.}
We apply models trained on MIMIC-III directly to the eICU Collaborative Research Database \citep{eicu}, whose ontologies differ and whose events are partly free text. \ours tokenizes both structured code descriptions and free-text event strings without manual vocabulary alignment.

\definecolor{lightgray}{gray}{0.9}
\begin{table*}[t]
\centering
\caption{EHR prediction performance on eICU, reported as mean (std) AUPRC across folds. Across the 14 tasks, \ours improves over MedTok (one-sided Wilcoxon signed-rank test, $p=6.1\times10^{-5}$). Patient-level bootstrap tests (5{,}000 resamples) show significant gains ($p<0.05$) on 12/14 tasks over MedTok and 14/14 tasks over Code Emb.}
\label{tab:eicu_results}
\setlength{\tabcolsep}{2.2pt}
\footnotesize
\begin{tabular}{lcccccc}
\toprule
Task
  & Code Emb & ClinVec & MedTok
  & $\text{MedTok}_{\text{text}}$-top5
  & $\text{MedTok}_{\text{text}}$-top1
  & \textbf{\ours} \\
\midrule
\multicolumn{7}{l}{\textit{Operational tasks}} \\
LOS $<$7
  & 0.933(0.003) & 0.939(0.002) & 0.934(0.001)
  & 0.941(0.026) & 0.939(0.001) & \textbf{0.942(0.004)} \\
ICU Mortality
  & 0.259(0.018) & 0.211(0.010) & 0.282(0.009)
  & 0.162(0.066) & 0.145(0.048) & \textbf{0.349(0.005)} \\

\rowcolor{lightgray}
Mean
  & 0.596 & 0.575 & 0.608
  & 0.552 & 0.542 & \textbf{0.645} \\
\midrule
\multicolumn{7}{l}{\textit{Subcohort mortality tasks}} \\
Acute Kidney Injury
  & 0.318(0.032) & 0.237(0.019) & 0.347(0.013)
  & 0.233(0.031) & 0.213(0.002) & \textbf{0.408(0.011)} \\
Acute MI
  & 0.241(0.032) & 0.191(0.015) & 0.227(0.049)
  & 0.110(0.011) & 0.119(0.006) & \textbf{0.325(0.024)} \\
Acute Pancreatitis
  & 0.110(0.034) & 0.121(0.015) & 0.140(0.064)
  & 0.073(0.015) & 0.077(0.000) & \textbf{0.178(0.049)} \\
Atrial Fibrillation
  & 0.200(0.027) & 0.169(0.018) & 0.214(0.019)
  & 0.125(0.020) & 0.095(0.002) & \textbf{0.290(0.012)} \\
ARF/ARDS
  & 0.285(0.025) & 0.253(0.014) & 0.363(0.022)
  & 0.250(0.034) & 0.254(0.004) & \textbf{0.397(0.020)} \\
Cirrhosis
  & 0.239(0.020) & 0.284(0.073) & 0.290(0.082)
  & 0.208(0.011) & 0.194(0.001) & \textbf{0.325(0.062)} \\
Heart Failure
  & 0.177(0.016) & 0.140(0.003) & 0.196(0.011)
  & 0.136(0.011) & 0.125(0.003) & \textbf{0.247(0.013)} \\
Liver Failure
  & 0.323(0.027) & 0.372(0.022) & 0.375(0.027)
  & 0.326(0.001) & 0.353(0.002) & \textbf{0.487(0.028)} \\
GI Bleeding
  & 0.153(0.027) & 0.164(0.009) & 0.226(0.013)
  & 0.134(0.038) & 0.132(0.005) & \textbf{0.331(0.025)} \\
Severe Sepsis/Septic Shock
  & 0.286(0.018) & 0.300(0.014) & 0.287(0.017)
  & 0.279(0.054) & 0.274(0.000) & \textbf{0.403(0.013)} \\
Cerebrovascular Disease
  & 0.136(0.027) & 0.147(0.033) & 0.105(0.011)
  & 0.123(0.021) & 0.104(0.000) & \textbf{0.294(0.010)} \\
Subarachnoid Hemorrhage
  & 0.191(0.027) & 0.199(0.016) & 0.189(0.021)
  & 0.222(0.043) & 0.199(0.004) & \textbf{0.401(0.099)} \\
\rowcolor{lightgray}
Mean
  & 0.222 & 0.215 & 0.247
  & 0.185 & 0.178 & \textbf{0.340} \\
\bottomrule
\end{tabular}
\vspace{-5pt}
\end{table*}

\textbf{Baselines.}
On BEL benchmarks, we compare with a $k$-means tokenizer on frozen Qwen3-Embedding-8B embeddings~\citep{qwen3embedding}, which clusters UMLS training embeddings into $k{=}100{,}000$ centroids and assigns each text to its nearest centroid, and with the text branch of MedTok~\citep{medtok}, a state-of-the-art medical tokenizer that learns a VQ codebook with separate text and knowledge-graph branches. The top-5 variant (MedTok's default) represents each input by its five nearest codes, and we treat each unique 5-code combination as a token ($\text{MedTok}_{\text{text}}$-top5). The top-1 variant uses only the nearest code ($\text{MedTok}_{\text{text}}$-top1), which makes it directly comparable to \ours. For EHR prediction, we additionally compare against Code Emb, which treats each medical code as a separate token, ClinVec~\citep{clinvec}, a pretrained knowledge graph embedding, and full MedTok. Under zero-shot transfer, the closed-vocabulary baselines (Code Emb, ClinVec, full MedTok) drop free text and codes outside the source vocabulary.

\textbf{Implementation details.}
We use frozen Qwen3-Embedding-8B~\citep{qwen3embedding} with precomputed instruction-conditioned embeddings, a codebook of 100,000 entries in a 512-dimensional VQ space, and four views per concept during training. Enriched descriptions are generated with MedGemma-1.5-4B, PPMI is built from a private health system of 7.1 million patients (a public alternative is evaluated in Section~\ref{sec:ablation}), and the relation graph uses UMLS medical relations. Once trained, \ours is applied offline to unique clinical strings, and the cached token IDs feed the downstream models (Appendix~\ref{vq_implementation}).
\begin{table*}[t]
\centering
\caption{\textbf{Direct open-vocabulary tokenization evaluation.}
Tests whether each tokenizer maps unseen mentions and their ground-truth concept
descriptions to one code while preserving purity.}
\label{tab:bel}
\footnotesize
\begin{tabular}{lccccccccc}
\toprule
 & \multicolumn{3}{c}{NCBI} & \multicolumn{3}{c}{MedMentions} & \multicolumn{3}{c}{COMETA} \\
\cmidrule(lr){2-4} \cmidrule(lr){5-7} \cmidrule(lr){8-10}
Model & Acc. & Pur. & F1 & Acc. & Pur. & F1 & Acc. & Pur. & F1 \\
\midrule
Qwen3-8B + $k$-means 
& 0.487 & 0.928 & 0.639 
& 0.450 & 0.543 & 0.492 
& 0.648 & 0.807 & 0.718 \\

MedTok$_{\mathrm{text}}$-top5 
& 0.195 & 0.986 & 0.326 
& 0.045 & 0.885 & 0.085 
& 0.231 & 0.913 & 0.368 \\

MedTok$_{\mathrm{text}}$-top1 
& 0.269 & 0.360 & 0.308 
& 0.131 & 0.082 & 0.101 
& 0.384 & 0.101 & 0.160 \\

\ours{} 
& 0.524 & 0.907 & \bfseries 0.664 
& 0.471 & 0.516 & \bfseries 0.493 
& 0.682 & 0.794 & \bfseries 0.734 \\
\bottomrule
\end{tabular}
\vspace{-10pt}
\end{table*}

\subsection{\ours is a discrete neural interlingua}
We first test whether \ours functions as a discrete neural interlingua, where synonymous surface forms should collapse to the same token, while each token should remain semantically coherent. As shown in Table~\ref{tab:bel}, \ours achieves the highest F1 on every dataset across domains.

The \ours codebook should also provide a meaningful embedding space for downstream EHR models. We assess this qualitatively by tokenizing concepts from major medical vocabularies, retrieving their assigned VQ code embeddings, and projecting them with UMAP \citep{umap}. As shown in Figure~\ref{fig:umap}, concepts cluster by ontology-derived categories that are not used as direct classification targets, including ICD disease chapters, LOINC measurement classes, SNOMED semantic groups, and procedure body-system categories. The resulting organization suggests that medical knowledge-guided training shapes the codebook into a clinically structured discrete space.

\subsection{\ours improves in-domain EHR modeling}
Table~\ref{tab:mimic_results} reports AUPRC on 20 MIMIC-III tasks spanning operations, subcohort mortality, and survival. \ours achieves the strongest mean performance in all three task families, improving over MedTok by 0.9, 4.6, and 4.6, respectively. Gains are most pronounced on clinically specific endpoints, including subarachnoid hemorrhage (+16.2), brain cancer (+10.0), cirrhosis (+8.6), and lymphoid leukemia (+6.6), suggesting that the shared semantic codebook improves statistical sharing across related codes. \ours also remains the best tokenizer with two stronger downstream backbones, GT-BEHRT~\citep{poulain2024gtbehrt} and TransformEHR~\citep{yang2023transformehr} (Appendix~\ref{app:architectures}).

\subsection{\ours improves zero-shot transfer to eICU}
\label{sec:eicu}
Table~\ref{tab:eicu_results} tests MIMIC-III-trained models directly on eICU. Of 3,578,036 eICU events, closed-vocabulary baselines retain only 466,408 (787 unique items), whereas \ours retains all of them, mapped to 4,102 distinct codes. \ours achieves the best transfer in both task families, improving over MedTok by 3.7 points on operational tasks and 9.3 points (38\% relative) on subcohort mortality, with large gains on subarachnoid hemorrhage (+21.2), cerebrovascular disease (+18.9), severe sepsis/septic shock (+11.6), and liver failure (+11.2). The open-vocabulary tokenizer is thus more robust to ontology shift and free-text documentation.

\textbf{Matched input coverage.} Part of this gain is from coverage, the intended benefit of an open-vocabulary tokenizer. To isolate representation quality, we restrict all models to codes available to closed-vocabulary baselines (Table~\ref{tab:eicu_matched_input}, Appendix~\ref{app:matched}). \ours remains the best on both task-family means, improving over MedTok by 1.0 (operational) and 1.9 (subcohort mortality). Coverage explains most of the full-input margin, and representation adds a consistent gain.

\subsection{\ours enables transferable interpretability}
\begin{figure}[t!]
  \centering
  \includegraphics[width=0.8\textwidth]{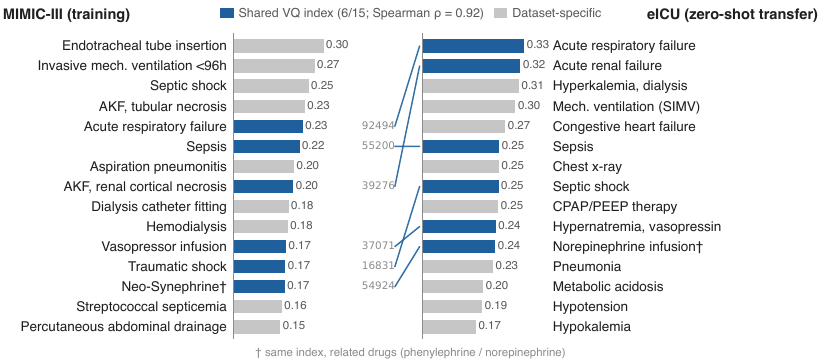}
  \caption{\textbf{Transferable attribution.} Top-15 tokens by IG attribution to cirrhosis mortality on MIMIC-III (left) and zero-shot eICU (right). Shared VQ indices are blue, others gray.}
  \label{fig:figure4}
  \vspace{-12pt}
\end{figure}

Because \ours maps structured codes and free text into a shared token space, attributions are comparable across datasets. Figure~\ref{fig:figure4} shows population-level integrated gradients (IG)~\citep{sundararajan2017axiomaticattributiondeepnetworks} for cirrhosis mortality trained on MIMIC-III and transferred to eICU. Across shared codes, attribution scores are strongly rank-correlated between datasets (Spearman $\rho=0.92$, $p=4\times10^{-26}$), so the importance profile transfers despite distribution shift. Six of the top 15 tokens appear in both datasets, against an expected 0.46 under a frequency-matched null over codes passing the attribution filter ($\geq$5\% of patients; permutation $p=5\times10^{-6}$), and overlap is also significant at top-10 and top-20. The shared tokens cover sepsis, shock, acute kidney and respiratory failure, and vasopressors, which match known mortality risk factors in decompensated cirrhosis. At $K{=}100$k, a token can also merge closely related concepts, such as the vasopressors phenylephrine and norepinephrine (token 54924), a granularity trade-off that code purity quantifies (Table~\ref{tab:bel}). These attributions describe predictive associations only.

\subsection{Ablation experiments}
\label{sec:ablation}
\begin{table}[t]
  \centering
  \begin{minipage}[t]{0.52\textwidth}
    \centering
    \caption{\textbf{Leave-one-out ablation.} Mean AUPRC $\pm$ std over all tasks.}
    \label{tab:ablation}
    \footnotesize
    \setlength{\tabcolsep}{4pt}
    \begin{tabular}{lcc}
    \toprule
    Variant & MIMIC-III & eICU \\
    \midrule
    $-\mathcal{L}_{\mathrm{agree}}$ & 0.554$\pm$0.003 & 0.378$\pm$0.008 \\
    $-\mathcal{L}_{\mathrm{co}}$ (PPMI) & 0.562$\pm$0.008 & 0.370$\pm$0.011 \\
    $-\mathcal{L}_{\mathrm{rel}}$ & 0.564$\pm$0.010 & 0.375$\pm$0.009 \\
    $-\mathcal{L}_{\mathrm{desc}}$ & 0.561$\pm$0.018 & 0.380$\pm$0.006 \\
    \midrule
    \ours{} full & \textbf{0.568$\pm$0.011} & \textbf{0.384$\pm$0.007} \\
    \bottomrule
    \end{tabular}
  \end{minipage}\hfill
  \begin{minipage}[t]{0.44\textwidth}
    \centering
    \caption{\textbf{PPMI teacher source.} Mean AUPRC over all tasks.}
    \label{tab:ppmi}
    \footnotesize
    \setlength{\tabcolsep}{4pt}
    \begin{tabular}{lcc}
    \toprule
    PPMI teacher & MIMIC-III & eICU \\
    \midrule
    Private (default) & \textbf{0.568} & \textbf{0.384} \\
    Public (COHD + GoM) & 0.565 & \textbf{0.384} \\
    None & 0.562 & 0.370 \\
    \bottomrule
    \end{tabular}
  \end{minipage}
  \vspace{-8pt}
\end{table}

Table~\ref{tab:ablation} reports leave-one-out ablations over all tasks, with per-family and additive results in Appendix~\ref{app:ablations}. Removing the agreement loss causes the largest overall drop, and removing PPMI a substantial drop, particularly on eICU. These two dominate downstream prediction and transfer, while enriched descriptions and relation-graph supervision give smaller, task-dependent gains. Against the mean of the four ablated variants, the full objective improves AUPRC by 0.008 (95\% CI [0.001, 0.014]; better on 24 of 34 tasks; Wilcoxon $p=0.011$) and has the best mean rank (Friedman $p=0.049$). In the additive ablation (Table~\ref{tab:ablation_additive}), the description and graph teachers mainly help in-domain, improving MIMIC-III operational AUPRC from 0.586 to 0.592 over the baseline with $\mathcal{L}_{\mathrm{agree}}$ and PPMI, while eICU results are comparable.

To separate the frozen encoder from the tokenizer objectives, we compare \ours with same-encoder controls that keep the Qwen3-Embedding-8B embeddings, BEHRT model, and input coverage fixed and form tokens by $k$-means on the 4096-dimensional embeddings or on 512-dimensional PCA projections matching our quantization space (Table~\ref{tab:encoder_tokenizer_ablation} in Appendix~\ref{app:ablations}, which also varies the frozen encoder). While $k$-means achieves performance close to \ours on MIMIC-III, both controls trail it on zero-shot eICU, where the downstream model cannot relearn task-specific token mappings. Strong text embeddings and full coverage thus drive in-domain performance, while the objectives of \ours add transfer robustness and open-vocabulary mapping quality.

Finally, we rebuild the PPMI teacher from public sources. We compute PPMI matrices from Columbia Open Health Data (COHD) concept co-occurrence counts~\citep{ta2018cohd} and from Stanford Graph of Medicine (GoM) co-occurrence statistics~\citep{finlayson2014graph}, map both to our concept groups, and average them (Appendix~\ref{vq_training_data}). The public teacher performs nearly identically to the private one (Table~\ref{tab:ppmi}), so the reported gains do not depend on private co-occurrence data.

\section{Conclusion}
Ontology-specific token interfaces hinder the transfer of EHR models across institutions and exclude much clinically relevant text. \ours reframes EHR harmonization as open-vocabulary discrete tokenization, mapping any clinical concept expressible as text into a shared VQ codebook. Across codebook analysis, entity linking, in-domain prediction, zero-shot transfer, and attribution, \ours provides a compact, reusable token interface for standard EHR Transformers, and on eICU it uses every event where closed-vocabulary tokenizers use 13\%. Open-vocabulary tokenization may also help other domains where many identifiers denote one entity, such as genes or financial securities.

\textbf{Limitations.} \ours{} inherits noise and omissions from its supervision sources, and a systematic error such as an incorrect synonym group affects every use of that concept, so clinically critical mappings should be validated before high-stakes use. The default co-occurrence teacher uses private data, though a public one performs nearly identically. Transfer gains partly reflect a coverage asymmetry, since \ours{} ingests events that closed-vocabulary baselines must discard. \ours{} tokenizes each concept description without context, so it cannot resolve short-form polysemy in ambiguous abbreviations, and it is trained on English sources only. Our downstream models are trained per task, and pretrained generative backbones and more hospitals and languages remain future work.

\bibliographystyle{plainnat}
\bibliography{4-reference}

\appendix
\setcounter{table}{0}
\setcounter{figure}{0}
\renewcommand{\thetable}{A.\arabic{table}}
\renewcommand{\thefigure}{A.\arabic{figure}}
\onecolumn

\section{Impact statement}

This work proposes a discrete, ontology-agnostic tokenizer that aims to improve the interoperability of EHR-based machine learning models across coding systems, versions, and institutions, and to enable more structured use of information extracted from clinical notes. If adopted in clinical research or deployed in downstream decision-support tools, such representations could improve portability and reduce the engineering burden of manual code mapping, potentially accelerating model development and evaluation across sites. However, the approach inherits common risks of EHR modeling. Biases and missingness in clinical data may be preserved, performance may vary across demographic groups or care settings, and errors in concept normalization could propagate into downstream predictions. In particular, a systematic error in a supervision source, such as an incorrectly curated synonym group, affects every downstream use of that concept, so mappings for clinically critical concepts should be independently validated before deployment in high-stakes settings.

\section{Data processing}

\subsection{Training data preparation}
\label{vq_training_data}
We construct the \ours pretraining corpus as a set of concept groups, each associated with multiple textual views. We begin from UMLS 2025AB MRCONSO and retain English, unsuppressed entries from the target clinical and biomedical vocabularies. An LLM (Qwen3.5-35B-A3B \citep{qwen3.5}) is used to filter the relations that are synonyms or medically related. We then merge CUIs connected by synonym-classified relations using union-find, where the resulting connected components define \texttt{CUI\_GROUP} identifiers. All retained source strings from CUIs in the same \texttt{CUI\_GROUP} are treated as positive textual views. We remove null and numeric-only strings, require at least two views per group before augmentation, and augment under-viewed groups so that each retained \texttt{CUI\_GROUP} has at least five distinct descriptions.

\textbf{LLM-generated text resources.}
For minimum-view augmentation, we use Qwen3.5-35B-A3B \citep{qwen3.5} to generate additional equivalent names for concept groups with fewer than five descriptions. The system prompt instructs the model to act as a medical expert and generate only new alternative names that are concise, medically accurate, semantically equivalent, of similar granularity, distinct from all existing names, and returned as a JSON array without explanations or metadata. The user prompt is as follows.
\begin{quote}\small
Representative concept name: \{representative name\}. Existing equivalent names already known: \{existing names\}. Generate \{n\} additional equivalent names that are distinct from all existing names. Output only a JSON array of strings.
\end{quote}
For description-teacher supervision, the pipeline uses a CUI-level table with columns \texttt{CUI}, \texttt{STR}, and \texttt{enriched\_description}. These enriched descriptions were generated offline with MedGemma-1.5-4B \citep{sellergren2026medgemma} and are aligned to \texttt{CUI\_GROUP}; we release the generated descriptions and prompts with our code. The system prompt is as follows.
\begin{quote} \small
    You are a medical expert. Write only plain-text responses. Never use markdown, headers, bullet points, numbered lists, or bold text. Do not include thinking steps or generation notes – only the final description. Keep it concise.
\end{quote}
and the user prompt is as follows.

\begin{quote}\small
    In plain-text sentences, describe what this concept is and its primary clinical relevance. Be informative and concise.
\end{quote}

\textbf{Embedding clinical strings and enriched descriptions.}
All tokenizer-input strings are embedded once with the frozen Qwen3-Embedding-8B encoder and cached before VQ training. We use instruction-aware formatting,
\texttt{Instruct:} \emph{Given a clinical concept string, produce an embedding that preserves exact biomedical concept identity across synonyms, ontology versions, abbreviations, and local hospital wording. Strings that refer to the same medical concept should be close. Strings that are clinically related but not equivalent should remain distinguishable.} \texttt{Query: \{text\}}.
For enriched descriptions, we use the analogous ``medical concept description'' instruction.

\textbf{Empirical co-occurrence resource.}
We build raw PPMI from claims-derived structured events from a private healthcare system containing 7.1 million patients. The event sources are problem-list diagnoses, laboratory events, procedures, medications, and immunizations. Raw codes are normalized and mapped to UMLS using ICD10CM for diagnoses, LOINC for laboratories, CPT for procedures, RxNorm for medications, and CVX for immunizations. The UMLS code map is then collapsed to \texttt{CUI\_GROUP}; codes that map ambiguously to multiple concept groups are excluded.

After mapping, we deduplicate events at the patient--date--\texttt{CUI\_GROUP} level and define contexts by patient and 30-day calendar time bin. Let $B\in\{0,1\}^{D\times |\mathcal{C}|}$ be the sparse context--concept matrix over $D$ patient-window contexts. We compute $C=B^\top B$, where $C_{ij}$ is the number of contexts containing both concepts $i$ and $j$, and $n_i=C_{ii}$ is the number of contexts containing concept $i$. The final sparse teacher is the positive pointwise mutual information (PPMI) matrix
\begin{equation}
M_{ij}=\max\left(0,\log_2 \frac{D\,C_{ij}}{n_i n_j}\right).
\end{equation}

\textbf{Public co-occurrence resource.}
To remove the dependence on private data, we also build a PPMI teacher from two public resources. Columbia Open Health Data (COHD)~\citep{ta2018cohd} provides concept prevalence and pairwise co-occurrence counts from the Columbia University Irving Medical Center EHR, encoded with OMOP standard concepts. The Stanford Graph of Medicine~\citep{finlayson2014graph} provides co-occurrence counts of UMLS concepts mined from clinical narratives. We map COHD concepts to UMLS through their source vocabularies, collapse both resources to \texttt{CUI\_GROUP}, compute a PPMI matrix from each resource's co-occurrence and marginal counts as above, and average the two matrices to form the public teacher.

\textbf{Medical relation knowledge graph.}
The medical relation knowledge graph $G_{\mathrm{rel}}$ is built from UMLS relations classified as synonymous or medically related by Qwen3.5-35B-A3B \citep{qwen3.5}, after mapping both endpoints to \texttt{CUI\_GROUP}. We remove self-loops after grouping, build an undirected graph over concept groups, and precompute two-hop neighborhoods for relation-based supervision during tokenizer training.

\section{Implementation details}

\subsection{\ours training}
\label{vq_implementation}
We initialize the codebook with $k$-means over the precomputed encoder embeddings projected into the quantization space, update codebook statistics with EMA, and periodically revive under-used codes from current encoder outputs. Post-quantization objectives use rotation-trick gradient transport. Semantic teachers from enriched descriptions, raw PPMI co-occurrence, and the medical relation knowledge graph $G_{\mathrm{rel}}$ are applied as separate weighted contrastive losses, and PPMI-neighbor sampling is used to increase the frequency of informative co-occurrence pairs in minibatches. Additional hyperparameters are listed in Table~\ref{tab:training_config}. Training requires 16GB. All experiments were trained on a single NVIDIA A100 GPU.

\ours is applied once, offline, as a vocabulary-level preprocessing step over unique clinical-concept strings. Each unique concept string is passed once through the model to obtain a discrete integer token id; these ids are then cached. At training and inference time the downstream EHR model consumes only these cached integer ids, and the text encoder is never invoked again. Concretely, tokenizing the entire concept vocabulary takes $\approx$70s per dataset on a single NVIDIA RTX PRO 6000 GPU. The 12,533 MIMIC-III concepts take 71.2s (5.7 ms/concept) and the 10,779 eICU concepts take 69.9s (6.5 ms/concept). Tokenization cost therefore scales only with the number of unique concepts and is independent of dataset size and sequence length. The text encoder and adaptor are never loaded by the downstream model. The codebook ($100{,}000\times512$; 51.2M parameters, $\approx$102\,MB in fp16) is used once to initialize the downstream embedding table (Appendix~\ref{behrt_implementation}).

\begin{table*}[t]
\centering

\begin{minipage}[t]{0.48\textwidth}
\centering
\captionof{table}{Hyperparameters used in final \ours tokenizer training.}
\label{tab:training_config}
\footnotesize
\setlength{\tabcolsep}{7pt}
\begin{tabular}{l c}
\toprule
Hyperparameter & Value \\
\midrule
Adaptor hidden dim. & 1024 \\
VQ embedding dim. & 512 \\
Codebook size & 100{,}000 \\
Views per concept & 4 \\
Batch size & 1536 \\
Training epochs & 100 \\
\midrule
Adaptor learning rate & $5\times10^{-5}$ \\
Other learning rate & $1\times10^{-4}$ \\
Weight decay & $1\times10^{-5}$ \\
Warmup ratio & 0.1 \\
Max gradient norm & 1.0 \\
\midrule
Contrastive temperature & 0.07 \\
Post-VQ temperature & 0.07 \\
Teacher temperature & 0.15 \\
EMA decay & 0.99 \\
EMA $\epsilon$ & $1\times10^{-5}$ \\
Revive threshold & 2.0 \\
Agreement warmup & 10 epochs \\
PPMI batch fraction & 0.083 \\
\midrule
$\lambda_{\mathrm{vq}}$ & 0.25 \\
$\lambda_{\mathrm{pre}}$ & 1.0 \\
$\lambda_{\mathrm{post}}$ & 1.5 \\
$\lambda_{\mathrm{agree}}$ & 1000.0 \\
$\lambda_{\mathrm{rel}}$ & 0.1 \\
$\lambda_{\mathrm{co}}$ & 0.1 \\
$\lambda_{\mathrm{desc}}$ & 0.1 \\
$\lambda_{\mathrm{rec}}$ & 0.01 \\
\bottomrule
\end{tabular}
\end{minipage}
\hfill
\begin{minipage}[t]{0.48\textwidth}
\centering
\captionof{table}{Hyperparameters used in downstream BEHRT-style EHR training.}
\label{tab:behrt_config}
\footnotesize
\setlength{\tabcolsep}{7pt}
\begin{tabular}{l c}
\toprule
Hyperparameter & Value \\
\midrule
Hidden size & 288 \\
Number of hidden layers & 6 \\
Number of attention heads & 12 \\
Intermediate size & 512 \\
Dropout probability & 0.1 \\
Segmentation vocab size & 3 \\
Age vocab size & 91 \\
Sex vocab size & 2 \\
\midrule
Batch size & 256 \\
Max epochs & 100 \\
Learning rate & $5\times10^{-4}$ \\
Warmup proportion & 0.1 \\
Weight decay & $5\times10^{-4}$ \\
Early-stop patience & 10 \\
\bottomrule
\end{tabular}
\end{minipage}

\end{table*}

\subsection{Downstream EHR model}
\label{behrt_implementation}
We train an end-to-end BEHRT \citep{li2020behrt} model for every single task and integrate the tokens for time, visit, and patient information (sex, age) as in the original paper. Hyperparameters are listed in Table~\ref{tab:behrt_config}.

For in-domain EHR prediction, the Code Emb baseline trains BEHRT with a standard learned embedding table over structured codes. For the full MedTok baseline, medical codes are passed to the original MedTok model and the resulting code embeddings are used as model inputs. ClinVec uses knowledge graph embeddings following the official setup~\citep{clinvec}. For text baselines, in $\text{MedTok}_{\text{text}}$-top5 and $\text{MedTok}_{\text{text}}$-top1 settings, the official description of each medical code is encoded using the MedTok text encoder, and the average of the top-5 returned embeddings or the top-1 embedding is used, respectively. For \ours, the input embedding table has one row per codebook entry ($100{,}000\times512$) and is initialized from the trained \ours codebook, followed by a linear projection from 512 to the 288-dimensional BEHRT hidden size. Both the embedding table and the projection are trainable during BEHRT training. Each event is represented by the projected embedding of its assigned VQ index. All models are trained with 5-fold data split.

For zero-shot transfer to eICU, we directly apply models trained on MIMIC-III without fine-tuning. All five fold-specific checkpoints are evaluated on eICU, and the final performance is reported as the average across folds. Code Emb, MedTok, and ClinVec must drop much of the text information, since eICU provides no official code mapping. To further ensure fair comparison, $\text{MedTok}_{\text{text}}$-top5, $\text{MedTok}_{\text{text}}$-top1 settings and \ours embed all texts and structured code descriptions. We allow eICU VQ indices that were not observed in MIMIC training; we therefore evaluate transfer with the full, unfiltered tokenizer output. Because the embedding table is initialized from the codebook, rows for indices never observed in MIMIC-III training receive no gradient updates and retain their codebook initialization (up to weight decay), so these indices are mapped through the learned projection to embeddings that reflect their position in the shared semantic space.

\subsection{Integrated gradients analysis}
We attribute predictions to \ours tokens using integrated gradients (IG)~\citep{sundararajan2017axiomaticattributiondeepnetworks} in embedding space.
Let a patient sequence be represented by a sequence of VQ indices, and let the model input be the corresponding sequence of VQ embeddings.
For IG, we treat each token embedding as the attribution unit and use the all-zero vector as the baseline embedding.
We compute IG with respect to the positive-class (mortality) logit.

From per-patient IG scores, we compute population-level importance for each token in three steps.
(i) We filter to tokens that appear in at least 5\% of patients to avoid unstable estimates for rare tokens.
(ii) Within each patient, if a token index appears multiple times, we apply max pooling over its IG scores so that each patient contributes at most one attribution value per token (capturing the strongest evidence for that concept in the trajectory).
(iii) We then average pooled IG scores across patients (without taking absolute values) to obtain directional population-level importance, where positive scores indicate risk-associated concepts and negative scores indicate protective concepts.

\section{Task definition}

The selection of downstream prediction tasks is guided by both clinical relevance and data availability across the MIMIC-III and eICU datasets. In particular, we prioritize tasks that are well-established in the clinical machine learning literature. For disease-specific mortality and cancer survival tasks, cohort selection is further constrained by the need to ensure sufficient statistical power. We ensure that all evaluated tasks are supported by sufficient data for reliable modeling and evaluation.

\subsection{MIMIC-III}
\label{mimic_task}
We use PyHealth 2.0 \citep{pyhealth} to construct the length-of-stay and readmission cohort. For other tasks,
we follow a standard MIMIC-III cohort-construction pipeline. After filtering for age $\geq$ 18, ICU stay structure, and task-specific eligibility, we sample one admission per patient, so each patient contributes a single admission trajectory and all splits are patient-disjoint. We evaluate all downstream models with five-fold cross-validation. In each run, one fold serves as the validation set, one as the test set, and the remaining three as the training set, and the checkpoint with the best validation AUPRC is used for testing. eICU is never used for training, hyperparameter tuning, or model selection. Subcohort mortality is defined to predict ICU mortality for patients diagnosed with that specific disease within that stay. Survival tasks predict 1-year mortality after ICU discharge for patients diagnosed with a specific cancer. Cohort statistics are reproduced in Table~\ref{tab:dataset_stats}.

ICU length of stay, ICU mortality, and sub-cohort mortality tasks are predicted using data available within the first 24 hours of ICU admission. ICU readmission and survival tasks are predicted at discharge for patients discharged alive. ICU readmission is defined as a subsequent ICU admission within 30 days. For survival tasks, the prediction time is ICU discharge; patients who die on or before the ICU discharge date are excluded, and the label is positive if the recorded date of death falls within 365 days after ICU discharge. Patients without a recorded date of death are treated as survivors; we do not explicitly censor patients with incomplete follow-up.

MIMIC-III diagnosis and procedure codes (\texttt{DIAGNOSES\_ICD} and \texttt{PROCEDURES\_ICD}) carry no timestamps and are assigned at the end of the hospital stay, so we timestamp them at discharge time. For every task, model inputs are restricted to events timestamped at or before the prediction time. 

Disease-specific mortality tasks are built from ICD-9 cohorts corresponding to 584 (acute kidney injury); 410 (acute myocardial infarction (MI)); 577.0 (acute pancreatitis); 427.31 (atrial fibrillation); 518.81 (acute respiratory failure/acute respiratory distress syndrome (ARF/ARDS)); 571.5, 571.6 (cirrhosis); 428 (heart failure); 570, 572.2, 572.8 (liver failure); 578 (gastrointestinal (GI) bleeding); 785.52, 995.92 (severe sepsis/septic shock); 430, 431, 433, 434, 436 (cerebrovascular disease, comprising subarachnoid and intracerebral hemorrhage, all subcodes of precerebral (433) and cerebral (434) artery occlusion with or without recorded infarction, and acute ill-defined cerebrovascular disease); and 430 (subarachnoid hemorrhage). Cancer survival tasks are built from ICD-9 codes 155 (liver cancer); 191 (brain cancer, i.e., malignant neoplasm of brain; histology not specified); 204 (lymphoid leukemia); 205 (myeloid leukemia); and 185 (prostate cancer). 

\begin{table}[t]
\caption{Dataset statistics for prediction tasks on MIMIC-III and eICU. Total Stays is the number of ICU stays in each task cohort (one per patient); Events is the number of positive labels.}
\label{tab:dataset_stats}
\vspace{0.5em}
\centering
\footnotesize
\setlength{\tabcolsep}{2.5pt}
\begin{tabular}{l l cc cc} %
\toprule
 & & \multicolumn{2}{c}{MIMIC-III} & \multicolumn{2}{c}{eICU} \\
\cmidrule(lr){3-4} \cmidrule(lr){5-6}
Group & Task & Total Stays & Events & Total Stays & Events \\
\midrule
\multirow{3}{*}{Hospital Operation}
 & LOS $<$ 7 days     & 40757 & 18931 & 83166   & 21818   \\
 & ICU Mortality & 31650  & 2954  & 83166   & 7003 \\
 & Readmission        & 9515  & 2613   & --      & --   \\

\midrule
\multirow{12}{*}{Subcohort ICU Mortality}
 & Acute Kidney Injury & 6757  & 1370 & 8309 & 1367 \\
 & Acute MI            & 3681  & 439  & 5671 & 414 \\
 & Acute Pancreatitis  & 627   & 84   & 322  & 19 \\
 & Atrial Fibrillation & 8179  & 984  & 7491 & 669 \\
 & ARF/ARDS                & 4732  & 1327 & 11170 & 2384 \\
 & Cirrhosis           & 755   & 133  & 497 & 92 \\
 & Heart Failure       & 7921  & 957  & 6806 & 708 \\
 & Liver Failure       & 1090  & 389  & 622 & 162 \\
 & GI Bleeding         & 1384  & 219  & 4984 & 403 \\
 & Severe Sepsis/Septic Shock        & 2468  & 818  & 6033 & 1279 \\
 & Cerebrovascular Disease              & 3014  & 525  & 7101 & 689 \\
 & Subarachnoid Hemorrhage & 509 & 100 & 497 & 90 \\

\midrule
\multirow{5}{*}{Cancer Survival}
 & Liver Cancer        & 245 & 120 & -- & -- \\
 & Prostate Cancer     & 180 & 44 & -- & -- \\
 & Brain Cancer        & 143 & 66 & -- & -- \\
 & Lymphoid Leukemia   & 116 & 50 & -- & -- \\
 & Myeloid Leukemia    & 122 & 67 & -- & -- \\

\bottomrule
\end{tabular}
\end{table}

\subsection{eICU}
\label{eicu_task}
We likewise follow the same eICU cohort-construction protocol for diagnoses, medications, admission drugs, past medical history, and treatments, again keeping one ICU stay per patient. Because eICU lacks post-discharge follow-up, we do not evaluate readmission and survival tasks. 

In addition to the ICD-9 label filtering rules applied to MIMIC-III, we define disease-specific subcohorts in eICU with the ICD-10 diagnosis codes N17 (acute kidney injury); I21, I22 (acute MI); K85 (acute pancreatitis); I48.0-I48.2, I48.91 (atrial fibrillation); J80 (ARF/ARDS); K74 (cirrhosis);  I50 (heart failure); K72 (liver failure); K92.0, K92.1, and K92.2 (GI bleeding); R65.20, R65.21 (severe sepsis/septic shock); I60, I61, I62, I63, I64 (cerebrovascular disease); and I60 (subarachnoid hemorrhage). Because ARDS-specific coding is unavailable in ICD-9, the MIMIC-III ARF/ARDS cohort is defined by acute respiratory failure (518.81), whereas the eICU cohort uses ARDS (J80), a narrower subset; transfer results for this task should be interpreted with this definitional difference in mind. Cohort details are in Table~\ref{tab:dataset_stats}. eICU records are passed through the tokenizer without restricting predictions to VQ indices observed in the source dataset.

\section{Codebook visualization}
\label{app:umap}
\begin{figure}[h]
  \centering
  \includegraphics[width=\textwidth]{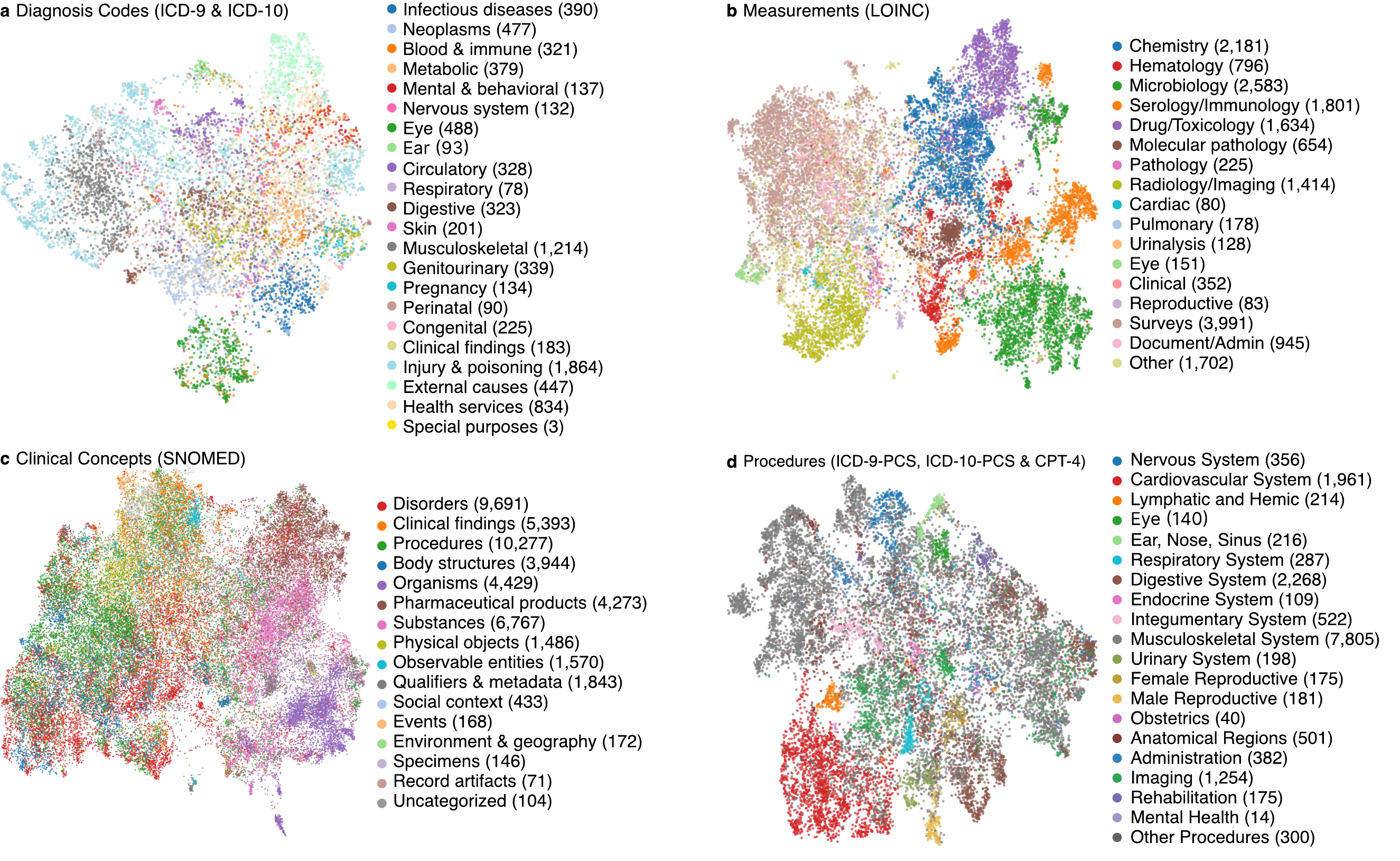}
  \caption{\textbf{UMAP visualization of the \ours codebook with full category legends.}
  Each point denotes a source concept represented by its assigned VQ code embedding and projected to two dimensions with UMAP. Colors indicate ontology-derived categories, with the number of concepts per category in parentheses, for (a) diagnosis codes (ICD-9 and ICD-10), (b) measurement concepts (LOINC), (c) clinical concepts (SNOMED CT), and (d) procedure codes (ICD-9-PCS, ICD-10-PCS, and CPT-4).}
  \label{fig:umap_full}
\end{figure}

Figure~\ref{fig:umap_full} shows the UMAP projections of Figure~\ref{fig:umap} with the complete category legends and the number of concepts in each category.

\section{Codebook size ablations}
\label{app:codebook}
\begin{figure}[t!]
  \vskip 0.2in
  \begin{center}
    \centerline{\includegraphics[width=\columnwidth]{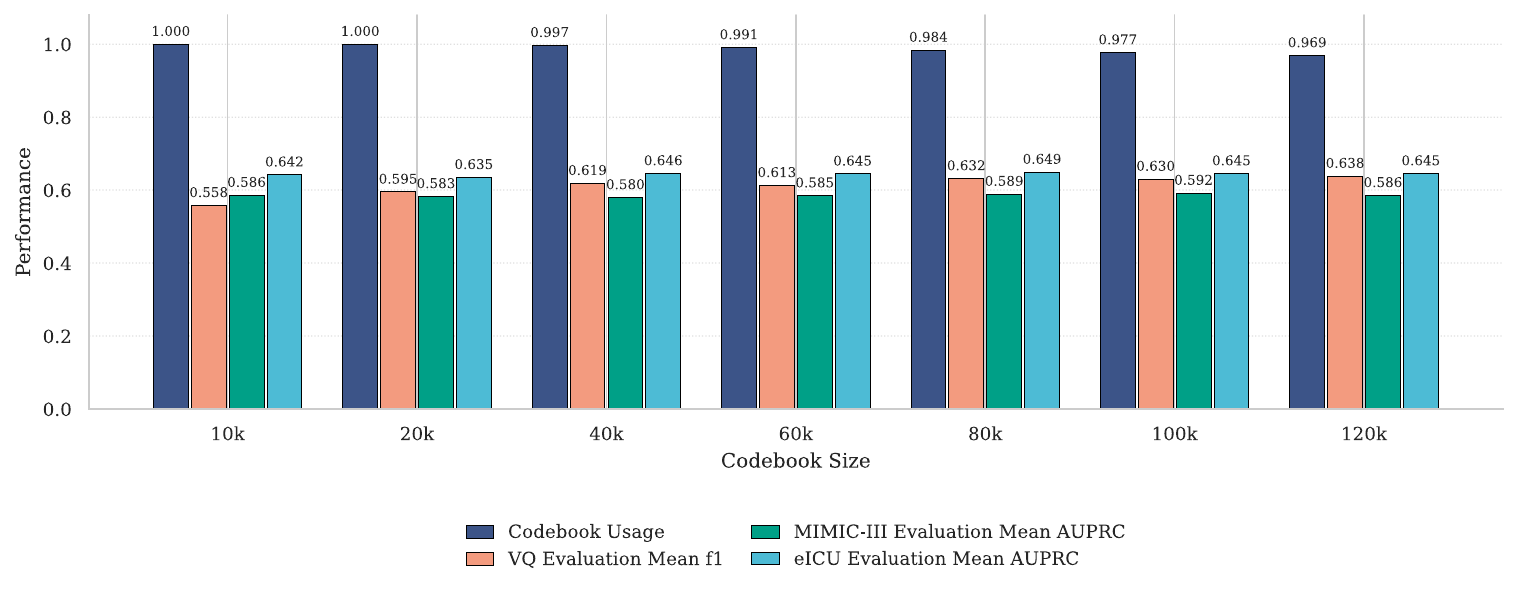}}
    \caption{\textbf{Effect of codebook size $K$.}
    We vary the VQ codebook size and report mean BEL (VQ Evaluation) F1 and operational-task mean AUPRC on MIMIC-III and zero-shot eICU. We additionally report codebook usage on the training corpus.}
    \vspace{-25pt}
    \label{fig:ablation}
  \end{center}
\end{figure}

\begin{table}[t]
\centering
\caption{\textbf{Biomedical entity linking F1 across codebook sizes.}}
\label{tab:codebook_bel}
\footnotesize
\begin{tabular}{lcccc}
\toprule
Codebook size & NCBI & MedMentions & COMETA & Mean \\
\midrule
10k  & 0.663 & 0.343 & 0.669 & 0.558 \\
20k  & 0.670 & 0.409 & 0.707 & 0.595 \\
40k  & 0.681 & 0.451 & 0.723 & 0.619 \\
60k  & 0.637 & 0.469 & 0.732 & 0.613 \\
80k  & 0.673 & 0.484 & 0.738 & 0.632 \\
100k & 0.664 & 0.493 & 0.734 & 0.630 \\
120k & 0.680 & 0.497 & 0.739 & 0.638 \\
\bottomrule
\end{tabular}
\end{table}

We vary the codebook size $K$ from 10k to 120k (Figure~\ref{fig:ablation}, Table~\ref{tab:codebook_bel}). Codebook size affects entity linking and downstream prediction differently. Entity-linking quality generally improves with $K$. Mean F1 rises from 0.558 at 10k to 0.638 at 120k, monotonically on MedMentions and nearly so on COMETA. Downstream AUPRC on operational tasks is comparatively flat, ranging from 0.580 to 0.592 on MIMIC-III and from 0.635 to 0.649 on eICU, with no improvement beyond 100k on either dataset. From 100k to 120k, MIMIC-III declines from 0.592 to 0.586 and eICU stays at 0.645. We therefore select $K{=}100$k as a cost-performance operating point that attains the best MIMIC-III value, lies within 0.004 of the best eICU value, and keeps entity-linking quality close to that of the largest codebook while keeping the vocabulary compact.
Across all $K$, codebook utilization on the training corpus stays above 96\%, indicating that \ours{}'s training stabilizes large codebooks and avoids the severe dead-code behavior commonly observed when scaling VQ models.

\section{Additional ablations}
\label{app:ablations}
This section reports the full component and encoder ablations summarized in Section~\ref{sec:ablation}, including leave-one-out results by task family (Table~\ref{tab:ablation_full}), the additive component ablation over operational tasks (Table~\ref{tab:ablation_additive}), and the encoder and same-encoder tokenizer controls (Table~\ref{tab:encoder_tokenizer_ablation}). Panel A of Table~\ref{tab:encoder_tokenizer_ablation} replaces the frozen Qwen3-Embedding-8B encoder with BioLORD-2023~\citep{biolord} or Qwen3-Embedding-0.6B while keeping the rest of the pipeline fixed. Both weaker encoders lower performance, although Qwen3-Embedding-0.6B retains most of the eICU transfer performance (0.372 vs.\ 0.384), indicating that the transfer gains come substantially from the learned mapping.
\begin{table*}[t]
\centering
\caption{\textbf{Leave-one-out component ablation by task family.}
Mean AUPRC $\pm$ standard deviation across folds for task-family
means on MIMIC-III and zero-shot eICU transfer.}

\label{tab:ablation_full}
\footnotesize
\setlength{\tabcolsep}{2pt}
\begin{tabular}{lccccccc}
\toprule
 & \multicolumn{4}{c}{MIMIC-III} & \multicolumn{3}{c}{eICU} \\
\cmidrule(lr){2-5} \cmidrule(lr){6-8}
Method & Oper. & Mort. & Surv. & All & Oper. & Mort. & All \\
\midrule
$-\mathcal{L}_{\mathrm{agree}}$ 
& 0.582{\scriptsize$\pm$0.010} 
& 0.524{\scriptsize$\pm$0.010} 
& 0.610{\scriptsize$\pm$0.023} 
& 0.554{\scriptsize$\pm$0.003} 
& 0.642{\scriptsize$\pm$0.006} 
& 0.334{\scriptsize$\pm$0.010} 
& 0.378{\scriptsize$\pm$0.008} \\

$-\mathcal{L}_{\mathrm{co}}$ (PPMI) 
& 0.585{\scriptsize$\pm$0.011} 
& 0.530{\scriptsize$\pm$0.013} 
& 0.628{\scriptsize$\pm$0.054} 
& 0.562{\scriptsize$\pm$0.008} 
& 0.644{\scriptsize$\pm$0.005} 
& 0.324{\scriptsize$\pm$0.013} 
& 0.370{\scriptsize$\pm$0.011} \\

$-\mathcal{L}_{\mathrm{rel}}$ 
& 0.590{\scriptsize$\pm$0.007} 
& \textbf{0.549}{\scriptsize$\pm$0.006} 
& 0.585{\scriptsize$\pm$0.045} 
& 0.564{\scriptsize$\pm$0.010} 
& 0.644{\scriptsize$\pm$0.005} 
& 0.330{\scriptsize$\pm$0.010} 
& 0.375{\scriptsize$\pm$0.009} \\

$-\mathcal{L}_{\mathrm{desc}}$ 
& 0.588{\scriptsize$\pm$0.009} 
& 0.537{\scriptsize$\pm$0.022} 
& 0.601{\scriptsize$\pm$0.051} 
& 0.561{\scriptsize$\pm$0.018} 
& 0.640{\scriptsize$\pm$0.008} 
& 0.337{\scriptsize$\pm$0.007} 
& 0.380{\scriptsize$\pm$0.006} \\

\midrule
\textbf{\ours{} full} 
& \textbf{0.592}{\scriptsize$\pm$0.007} 
& 0.533{\scriptsize$\pm$0.012} 
& \textbf{0.639}{\scriptsize$\pm$0.037} 
& \textbf{0.568}{\scriptsize$\pm$0.011} 
& \textbf{0.645}{\scriptsize$\pm$0.003} 
& \textbf{0.340}{\scriptsize$\pm$0.009} 
& \textbf{0.384}{\scriptsize$\pm$0.007} \\
\bottomrule
\end{tabular}

\end{table*}

\begin{table}[t]
\centering
\caption{\textbf{Additive component ablation.}
Mean AUPRC over operational tasks on MIMIC-III and zero-shot eICU transfer,
adding components to a baseline trained with only the equivalence, VQ, and
reconstruction losses.}
\label{tab:ablation_additive}
\footnotesize
\begin{tabular}{lcc}
\toprule
Objective & MIMIC-III Oper. & eICU Oper. \\
\midrule
Baseline & 0.585 & 0.641 \\
Baseline $+\,\mathcal{L}_{\mathrm{desc}}$ & 0.586 & 0.642 \\
Baseline $+\,\mathcal{L}_{\mathrm{co}}$ (PPMI) & 0.580 & 0.645 \\
Baseline $+\,\mathcal{L}_{\mathrm{rel}}$ & 0.584 & 0.635 \\
Baseline $+\,\mathcal{L}_{\mathrm{agree}}$ & 0.583 & 0.644 \\
Baseline $+\,\mathcal{L}_{\mathrm{agree}}+\mathcal{L}_{\mathrm{desc}}$ & 0.584 & 0.637 \\
Baseline $+\,\mathcal{L}_{\mathrm{agree}}+\mathcal{L}_{\mathrm{co}}$ (PPMI) & 0.586 & \textbf{0.651} \\
Baseline $+\,\mathcal{L}_{\mathrm{agree}}+\mathcal{L}_{\mathrm{rel}}$ & 0.584 & 0.641 \\
\midrule
\ours{} full & \textbf{0.592} & 0.645 \\
\bottomrule
\end{tabular}
\end{table}

\begin{table}[t]
\centering
\caption{\textbf{Encoder and tokenizer ablations.}
Panel A varies the frozen text encoder used by \ours{}. Panel B keeps the
same frozen Qwen3-Embedding-8B encoder fixed and compares \ours{} against
direct clustering controls. All results report mean AUPRC on 20 MIMIC-III
tasks and 14 zero-shot eICU transfer tasks.}
\label{tab:encoder_tokenizer_ablation}
\footnotesize
\begin{tabular}{llccc}
\toprule
Analysis & Variant & Dim. & MIMIC-III & eICU \\
\midrule
\multicolumn{5}{l}{\textbf{A. Alternative frozen encoders for \ours{}}} \\
Encoder & BioLORD-2023 & 768 & 0.551 & 0.312 \\
Encoder & Qwen3-Embedding-0.6B & 1024 & 0.550 & 0.372 \\
Encoder & Qwen3-Embedding-8B & 4096 & \bfseries 0.568 & \bfseries 0.384 \\
\midrule
\multicolumn{5}{l}{\textbf{B. Same-encoder clustering controls}} \\
Tokenizer & Qwen embeddings + $k$-means & 4096 & \bfseries 0.568 & 0.378 \\
Tokenizer & Qwen embeddings $\rightarrow$ PCA $\rightarrow$ $k$-means & 512 & 0.558 & 0.379 \\
Tokenizer & \ours{} & 512 & \bfseries 0.568 & \bfseries 0.384 \\
\bottomrule
\end{tabular}
\end{table}

\section{Additional EHR architectures}
\label{app:architectures}

To test whether the benefit of \ours{} depends on the specific downstream
BEHRT-style architecture used in the main experiments, we evaluated the same
tokenizer variants with two stronger EHR backbones, GT-BEHRT~\citep{poulain2024gtbehrt} and TransformEHR~\citep{yang2023transformehr}.
For each architecture, the tokenizer is the only component varied, while the
downstream training and evaluation protocol follows the main MIMIC-III
in-domain and MIMIC-to-eICU zero-shot transfer setup. As shown in
Table~\ref{tab:stronger_architectures}, \ours{} remains the best tokenizer
under both architectures and on both evaluation settings. This suggests that
the gains from open-vocabulary discrete tokenization are not specific to the
BEHRT-style model used in the main paper, but transfer to stronger EHR sequence
modeling backbones.
\begin{table}[t]
\centering
\caption{\textbf{Tokenizer comparison under stronger EHR architectures.}
We evaluate \ours{} as the token interface for GT-BEHRT and TransformEHR.
Results are mean AUPRC on 20 MIMIC-III tasks and 14 zero-shot eICU transfer tasks.}
\label{tab:stronger_architectures}
\footnotesize
\begin{tabular}{lcc}
\toprule
Tokenizer 
& MIMIC-III 
& eICU \\
\midrule
\multicolumn{3}{l}{\textbf{GT-BEHRT}} \\
Code Emb              & 0.572 & 0.313 \\
ClinVec               & 0.535 & 0.280 \\
MedTok                & 0.573 & 0.313 \\
MedTok$_{\mathrm{text}}$-top1 & 0.511 & 0.219 \\
MedTok$_{\mathrm{text}}$-top5 & 0.534 & 0.224 \\
\ours{}               & \bfseries 0.602 & \bfseries 0.414 \\
\midrule
\multicolumn{3}{l}{\textbf{TransformEHR}} \\
Code Emb              & 0.470 & 0.272 \\
ClinVec               & 0.481 & 0.264 \\
MedTok                & 0.512 & 0.299 \\
MedTok$_{\mathrm{text}}$-top1 & 0.434 & 0.227 \\
MedTok$_{\mathrm{text}}$-top5 & 0.499 & 0.218 \\
\ours{}               & \bfseries 0.553 & \bfseries 0.384 \\
\bottomrule
\end{tabular}
\end{table}
For GT-BEHRT, MIMIC-III and eICU in our benchmark contain a single admission
trajectory per patient, so the original multi-visit, per-visit graph
construction does not directly apply. We therefore use a single-visit
adaptation with a per-patient star topology, where a learned readout token
attends to and from all event tokens through dense masked attention, and edge
typing is collapsed to code-category embeddings. This preserves the
graph-attention mechanism of GT-BEHRT while matching the single-admission
structure of our benchmark.

\section{Zero-shot comparison on the mappable subset}
\label{app:matched}
\begin{table*}[t]
\centering
\caption{\textbf{Matched-input zero-shot transfer on eICU.}
All models are evaluated on the overlapping subset of eICU codes available to
closed-vocabulary baselines. We report mean AUPRC with standard deviation
across folds. Best results are bolded and second-best results are underlined.}
\label{tab:eicu_matched_input}
\footnotesize
\setlength{\tabcolsep}{2.2pt}
\begin{tabular}{lcccccc}
\toprule
Task & Code Emb & ClinVec & MedTok & $\text{MedTok}_{\text{text}}$-top5 & $\text{MedTok}_{\text{text}}$-top1 & \ours{} \\
\midrule
\multicolumn{7}{l}{\textbf{\textit{Operational tasks}}} \\
LOS $<$7
& 0.933(0.003)
& \textbf{0.939(0.002)}
& 0.934(0.001)
& 0.937(0.004)
& 0.926(0.003)
& \underline{0.938(0.002)} \\
ICU Mortality
& 0.259(0.018)
& 0.211(0.010)
& \underline{0.282(0.009)}
& 0.138(0.022)
& 0.130(0.010)
& \textbf{0.298(0.021)} \\
\textbf{Mean}
& 0.596
& 0.575
& \underline{0.608}
& 0.537
& 0.528
& \textbf{0.618} \\
\midrule
\multicolumn{7}{l}{\textbf{\textit{Subcohort mortality tasks}}} \\
Acute Kidney Injury
& 0.318(0.032)
& 0.237(0.019)
& \textbf{0.347(0.013)}
& 0.184(0.023)
& 0.185(0.011)
& \underline{0.345(0.010)} \\
Acute MI
& \textbf{0.241(0.032)}
& 0.191(0.015)
& 0.227(0.049)
& 0.102(0.020)
& 0.108(0.010)
& \underline{0.227(0.011)} \\
Acute Pancreatitis
& 0.110(0.034)
& 0.121(0.015)
& \underline{0.140(0.064)}
& 0.087(0.022)
& 0.089(0.025)
& \textbf{0.162(0.042)} \\
Atrial Fibrillation
& 0.200(0.027)
& 0.169(0.018)
& \underline{0.214(0.019)}
& 0.132(0.043)
& 0.084(0.009)
& \textbf{0.242(0.015)} \\
ARF/ARDS
& 0.285(0.025)
& 0.253(0.014)
& \textbf{0.363(0.022)}
& 0.231(0.014)
& 0.228(0.011)
& \underline{0.340(0.022)} \\
Cirrhosis
& 0.239(0.020)
& 0.284(0.073)
& \underline{0.290(0.082)}
& 0.196(0.020)
& 0.194(0.018)
& \textbf{0.301(0.050)} \\
Heart Failure
& 0.177(0.016)
& 0.140(0.003)
& \underline{0.196(0.011)}
& 0.130(0.016)
& 0.112(0.008)
& \textbf{0.201(0.006)} \\
Liver Failure
& 0.323(0.027)
& \underline{0.372(0.022)}
& \textbf{0.375(0.027)}
& 0.272(0.046)
& 0.302(0.047)
& 0.357(0.048) \\
GI Bleeding
& 0.153(0.027)
& 0.164(0.009)
& \underline{0.226(0.013)}
& 0.108(0.015)
& 0.125(0.006)
& \textbf{0.268(0.021)} \\
Severe Sepsis/Septic Shock
& 0.286(0.018)
& \textbf{0.300(0.014)}
& 0.287(0.017)
& 0.227(0.035)
& 0.230(0.015)
& \underline{0.299(0.017)} \\
Cerebrovascular Disease
& 0.136(0.027)
& \underline{0.147(0.033)}
& 0.105(0.011)
& 0.117(0.013)
& 0.104(0.010)
& \textbf{0.185(0.021)} \\
Subarachnoid Hemorrhage
& 0.191(0.027)
& \underline{0.199(0.016)}
& 0.189(0.021)
& 0.188(0.018)
& 0.183(0.022)
& \textbf{0.268(0.034)} \\
\textbf{Mean}
& 0.222
& 0.215
& \underline{0.247}
& 0.165
& 0.162
& \textbf{0.266} \\
\bottomrule
\end{tabular}
\end{table*}

Table~\ref{tab:eicu_matched_input} shows that matching coverage removes most, but not all, of the performance gap. Against the strongest baseline (MedTok), \ours's operational-mean margin falls from 3.7 to 1.0 points, and its subcohort-mortality margin falls from 9.3 to 1.9 points. \ours remains the best model on both task-family means, while MedTok leads the mean on neither.

\section{Worked example}
\label{app:worked_example}
Table~\ref{tab:worked_example} traces three surface forms of one concept and one distinct concept through the \ours pipeline of text encoding, projection, and nearest-code assignment.
\begin{table}[h]
\centering
\caption{\textbf{Worked example of \ours tokenization.} Each string is encoded by the frozen text encoder $f_{\eta}$, projected into the quantization space by $p_{\theta}$, and assigned to its nearest codebook entry. Three surface forms of dyspnea, from a clinical phrase, a colloquial expression, and the SNOMED CT term, are assigned the same code $i$, whereas the distinct concept diabetes mellitus is assigned a different code $j$. Because the codebook is fixed after training, strings from new code systems are tokenized in the same way without retraining.}
\label{tab:worked_example}
\footnotesize
\begin{tabular}{lccc}
\toprule
Input string $x$ & $h=f_{\eta}(x)$ & $z=p_{\theta}(h)$ & $k^\star(z)$ \\
\midrule
``shortness of breath'' (clinical phrase) & $h_1$ & $z_1$ & $i$ \\
``can't breathe'' (colloquial) & $h_2$ & $z_2$ & $i$ \\
``dyspnea'' (SNOMED CT term) & $h_3$ & $z_3$ & $i$ \\
\midrule
``diabetes mellitus'' (distinct concept) & $h_4$ & $z_4$ & $j\neq i$ \\
\bottomrule
\end{tabular}
\end{table}

\clearpage
\newpage
\end{document}